%% file: acl.tex
\documentclass[11pt]{article}

\usepackage[final]{acl}

\usepackage{times}
\usepackage{latexsym}
\usepackage{algorithm}
\usepackage{algorithmic}
\renewcommand{\algorithmicrequire}{\textbf{Input:}}
\renewcommand{\algorithmicensure}{\textbf{Output:}}
\usepackage[T1]{fontenc}

\usepackage[utf8]{inputenc}

\usepackage{microtype}

\usepackage{inconsolata}

\usepackage{graphicx}
\usepackage{float}
\usepackage{siunitx}
\usepackage{amssymb}
\usepackage{amsmath}
\usepackage{booktabs}
\usepackage{makecell}
\usepackage{subfigure} 
\usepackage{array} 
\usepackage{multirow}
\usepackage{bbding}
\usepackage{bm}
\usepackage{amssymb}
\usepackage{pifont}
\usepackage{url}
\usepackage{amsthm}

\title{Collaboration of Fusion and Independence: Hypercomplex-driven Robust Multi-Modal Knowledge Graph Completion}

\author{Zhiqiang Liu$^{1,3}$, Yichi Zhang$^{2,3}$, Mengshu Sun$^4$, Lei Liang$^4$, Wen Zhang$^{1,3}$\thanks{Corresponding Author.}
\\
$^1$School of Software Technology, Zhejiang University \\
$^2$College of Computer Science and Technology, Zhejiang University\\
$^3$ZJU-Ant Group Joint Lab of Knowledge Graph \\
$^4$Ant Group \\
\texttt{\{zhiqiangliu,zhang.wen\}@zju.edu.cn} \\}

\begin{document}
\maketitle
\begin{abstract}
Multi-modal knowledge graph completion (MMKGC) aims to discover missing facts in multi-modal knowledge graphs (MMKGs) by leveraging both structural relationships and diverse modality information of entities. Existing MMKGC methods follow two multi-modal paradigms: fusion-based and ensemble-based. Fusion-based methods employ fixed fusion strategies, which inevitably leads to the loss of modality-specific information and a lack of flexibility to adapt to varying modality relevance across contexts. In contrast, ensemble-based methods retain modality independence through dedicated sub-models but struggle to capture the nuanced, context-dependent semantic interplay between modalities. To overcome these dual limitations, we propose a novel MMKGC method \textbf{M-Hyper}, which achieves the coexistence and collaboration of fused and independent modality representations. Our method integrates the strengths of both paradigms, enabling effective cross-modal interactions while maintaining modality-specific information. Inspired by ``quaternion'' algebra, we utilize its four orthogonal bases to represent multiple independent modalities and employ the Hamilton product to efficiently model pair-wise interactions among them. Specifically, we introduce a Fine-grained Entity Representation Factorization (FERF) module and a Robust Relation-aware Modality Fusion (R2MF) module to obtain robust representations for three independent modalities and one fused modality. The resulting four modality representations are then mapped to the four orthogonal bases of a biquaternion for comprehensive modality interaction. Extensive experiments indicate its state-of-the-art performance with better robustness. Our dataset and code are available at \url{https://github.com/zjukg/M-Hyper}. 
\end{abstract}

\section{Introduction}
Multi-modal Knowledge Graphs (MMKGs)~\cite{MMKG} expand traditional knowledge graphs by incorporating additional multi-modal information, making them more powerful tools~\cite{MMKG-survey} for knowledge representation. This makes MMKGs valuable for various applications, including recommendation systems~\cite{KGAT} and natural language processing~\cite{TeleBERT,ontotune}. However, like traditional uni-modal knowledge graphs~\cite{unihr}, MMKGs also suffer from incomplete information~\cite{IKRL}; this limitation has been ameliorated through Multi-Modal Knowledge Graph Completion (MMKGC) methods.

\begin{figure}
\centering
\includegraphics[scale=0.6]{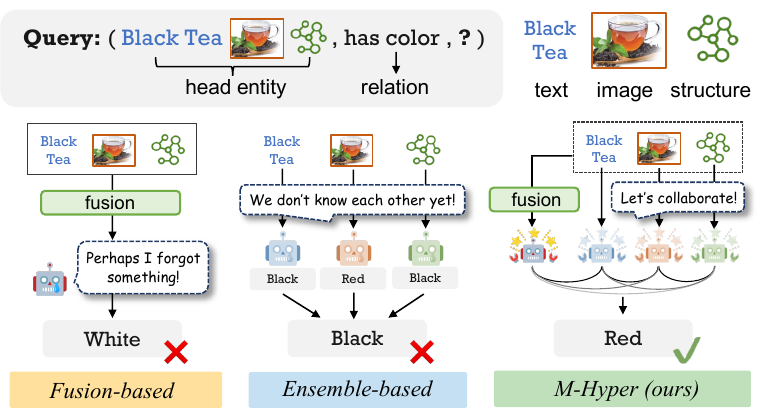}
\vspace{-4mm}
\caption{A simple example illustrates the difference between M-Hyper and existing paradigms.}
\vspace{-1mm}
\label{intro}
\end{figure}

As shown in Figure~\ref{intro}, existing MMKGC approaches fall into two paradigms: fusion-based and ensemble-based. Fusion-based methods~\cite{mygo} achieve cross-modality interaction via explicit fusion modules or dedicated cross-modality loss functions. Yet, their reliance on fixed fusion strategies often leads to suboptimal representation: crucial unique modality cues can be lost during fusion, and the model struggles to flexibly adapt to varying modality salience and synergies required in distinct reasoning contexts. Conversely, ensemble-based methods~\cite{IMF} preserve modality-specific characteristics by employing independent sub-models, but inevitably fail to capture subtle inter-modal dependencies and interactions that are critical for complex reasoning scenarios. This highlights a fundamental challenge: the modality requirements in MMKGs exhibit dynamic, context-dependent, and task-specific contributions, making rigid adherence to either independent or fully fused paradigms a significant limitation to the expressive power and adaptability of MMKGC models. Hence, we propose the following research question: is it possible to develop a method that \textbf{combines the strengths of both paradigms, adapting to both fused and independent modality requirements while dynamically enabling comprehensive cross-modal interactions?}

To address these limitations, we introduce \textbf{M-Hyper}, the first method to model MMKGs in a \textbf{\underline{hyper}}complex space. Inspired by quaternion algebra, where the four orthogonal basis elements preserve linear independence, M-Hyper explicitly separates distinct modality representations to retain original modal information and leverages the Hamilton product to facilitate comprehensive pairwise interactions among modalities. To enhance the robustness of modality representations, we design two novel modules: Fine-grained Entity Representation Factorization (FERF), which yields robust representations for three independent modalities, and Robust Relation-aware Modality Fusion (R2MF), which produces one robust fused modality representation. These four representations are mapped to the four orthogonal bases of a biquaternion, and a biquaternion-based scoring function is used to fully capture cross-modal semantic information.
Experimental results show that our M-Hyper achieves state-of-the-art performance on three MMKGC datasets and exhibits high robustness and computational efficiency. Our contributions can be summarized as follows:
\begin{itemize}

\item We highlight the limitations of existing MMKGC paradigms and propose a novel biquaternion-based representation approach that simultaneously preserves both individual and fused modalities.
\item We propose M-Hyper, the first MMKGC method operating in a hypercomplex (biquaternion) space, enabling robust coexistence and collaboration of fused and independent modality representations.
\item Extensive empirical evaluation on three MMKGC benchmarks demonstrates that M-Hyper outperforms 18 existing baseline methods, exhibiting superior robustness and computational efficiency.
\end{itemize}

\section{Related Works}
\subsection{Hypercomplex-based KG Embedding}
Knowledge graph embedding (KGE) aims to project entities and relations into continuous vector spaces to capture complex relational patterns. Classic KGE methods include translational models (e.g, TransE~\cite{TransE}) and semantic-matching models (e.g., ComplEx~\cite{complEx}). To enhance representation capability~\cite{pami}, hypercomplex spaces have been introduced: QuatE~\cite{quate} first extends embeddings to quaternion space, improving the modeling of symmetry and hierarchy. Subsequently, DualE~\cite{duale} and BiQUE~\cite{bique} further generalize to dual quaternions and biquaternion spaces, supporting richer relational composition via translation and rotation. Hypercomplex representations exhibit strong expressiveness for hierarchical, symmetric, and complex relational structures, and have recently been applied to more advanced KGC scenarios~\cite{bive}. However, prior hypercomplex-based methods focus only on uni-modal knowledge graphs, and their potential for handling rich multi-modal semantics remains underexplored. In contrast, our approach is the first to leverage biquaternion space for MMKGs, supporting both multi-modality and complex relational transformations.

\subsection{Multi-modal Knowledge Graph Completion}
Existing Multi-modal Knowledge Graph Completion (MMKGC) methods extend traditional KGC models by integrating various modalities (e.g., structural information in MMKG, as well as image and textual information of entities). From the perspective of multi-modality modeling, current MMKGC methods can be categorized into multi-modal fusion-based methods and multi-modal ensemble-based methods. 

Multi-modal fusion-based methods aim to design sophisticated multi-modal fusion modules to achieve modality alignment. Earlier modality fusion methods like IKRL~\cite{IKRL} and TransAE~\cite{TransAE} achieve efficient modality fusion by introducing cross-modal loss functions, demonstrating the effectiveness of cross-modal interactions. Furthermore, research community continues to propose more complex modality fusion designs with advanced techniques, such as OTKGE~\cite{OTKGE} with optimal transfer, AdaMF~\cite{AdaMF-MAT} with adversarial training and MyGO~\cite{mygo} with fine-grained multi-modal tokenization. However, these modal fusion methods rarely preserve independent modalities and excessively rely on fixed fusion strategies. Therefore, this paradigm inevitably introduces information loss during the modality fusion stage and makes it difficult to adapt to the flexible modality requirements during the reasoning stage.  

In contrast, classic modality ensemble methods like MoSE~\cite{MOSE} usually design individual sub-models for different modalities, and the individual representations obtained by these sub-models are integrated for joint decision-making. Subsequently, IMF~\cite{IMF} utilizes tensor decomposition to fuse multi-modality information and introduces a sub-model of joint modalities into the modality ensemble method. We consider this a promising beginning for achieving joint decision-making that incorporates both fused and independent modalities. After that, MoMoK~\cite{momok} follows this idea and decouples the modal representations through the MoE network with minimizing their mutual information. However, under the multi-modality ensemble paradigm, the sub-models lack explicit mechanisms for comprehensive cross-modal interaction, thereby limiting their overall modeling capability. 

\section{Preliminaries}
\textbf{Quaternion} number system was first proposed by~\citet{quaternion} to extend the complex numbers. The algebraic representation of a quaternion is typically expressed as:
\begin{equation}
    Q=a\mathbf{1} + b\mathbf{i} + c\mathbf{j} + d\mathbf{k},
\end{equation}
where the coefficient $a$ is a real number representing real part, the coefficients $b,c,d$ are real numbers representing imaginary part, and $ \mathbf{1}, \mathbf{i}, \mathbf{j}, \mathbf{k} $ are the orthogonal basis vectors or basis elements, which satisfy the following multiplication properties: $\mathbf{i}1=1\mathbf{i}=\mathbf{i}$, $\mathbf{j}1=1\mathbf{j}=\mathbf{j}$, $\mathbf{k}1=1\mathbf{k}=\mathbf{k}$, $\mathbf{i}^2 = \mathbf{j}^2 = \mathbf{k}^2 = -1$, $\mathbf{ij}=-\mathbf{ji}=\mathbf{k}$, $\mathbf{jk} =-\mathbf{kj}= \mathbf{i}$, $\mathbf{ki} =-\mathbf{ik}= \mathbf{j}$, and $\mathbf{ijk}=-1$. 

\vspace{1mm}
\noindent \textbf{Hamilton Product} can be regarded as ``Quaternion Multiplication'', which is composed of all standard multiplications of factors in quaternions, defined as:
\begin{equation}
\begin{aligned}
\label{product}
    Q_1 \otimes Q_2
& =(a_1a_2-b_1b_2-c_1c_2-d_1d_2) \\
&\, +(a_1b_2+b_1a_2+c_1d_2-d_1c_2)\mathbf{i} \\
&\, +(a_1c_2-b_1d_2+c_1a_2+d_1b_2)\mathbf{j}  \\
&\, +(a_1d_2+b_1c_2-c_1b_2+d_1a_2)\mathbf{k}. 
\end{aligned}
\end{equation}

\noindent\textbf{Biquaternions} further extend quaternions, and their algebra can be considered as a tensor product $\mathbb{C} \otimes_\mathbb{R} \mathbb{H}$, where $\mathbb{C}$ is the field of complex numbers and $\mathbb{H}$ is the division algebra of (real) quaternions. Biquaternions extend the coefficients of quaternions to complex numbers, denoted as:
\begin{equation}
    Q=(a_r+a_i\mathbf{I}) + (b_r+b_i\mathbf{I})\mathbf{i} + (c_r+c_i\mathbf{I})\mathbf{j} + (d_r+d_i\mathbf{I})\mathbf{k},
\end{equation}
where $\mathbf{I}$ is the imaginary unit of the complex number field $\mathbb{C}$, satisfying $\textbf{I}^2=-1$. The algebra $\mathbb{C} \otimes_\mathbb{R} \mathbb{H}$ satisfies the commutation relations $\textbf{Ii}=\textbf{iI}$, $\textbf{Ij}=\textbf{jI}$, $\textbf{Ik}=\textbf{kI}$. 

\vspace{2mm}
\noindent\textbf{Hamilton Product of Biquaternions} can be seen as an extension of the Hamilton product of quaternions. Similarly, for two biquaternions $Q_1=a_1 + b_1\mathbf{i} + c_1\mathbf{j} + d_1\mathbf{k}=(a_{\text{r},1}+a_{\text{i},1}\mathbf{I}) + (b_{\text{r},1}+b_{\text{i},1}\mathbf{I})\mathbf{i} + (c_{\text{r},1}+c_{\text{i},1}\mathbf{I})\mathbf{j} + (d_{\text{r},1}+d_{\text{i},1}\mathbf{I})\mathbf{k}$ and $Q_2=a_2 + b_2\mathbf{i} + c_2\mathbf{j} + d_2\mathbf{k}=(a_{\text{r},2}+a_{\text{i},2}\mathbf{I}) + (b_{\text{r},2}+b_{\text{i},2}\mathbf{I})\mathbf{i} + (c_{\text{r},2}+c_{\text{i},2}\mathbf{I})\mathbf{j} + (d_{\text{r},2}+d_{\text{i},2}\mathbf{I})\mathbf{k}$, the multiplication is performed exactly as in Equation~\ref{product} for quaternions, but with all coefficients treated as complex numbers (with $\mathbf{I}^2=-1$). That is, the Hamilton product is defined in the same way, with addition and multiplication of coefficients carried out in the field of complex numbers $\mathbb{C}$. 

\begin{figure*}[ht]
\centering
\includegraphics[scale=0.557]{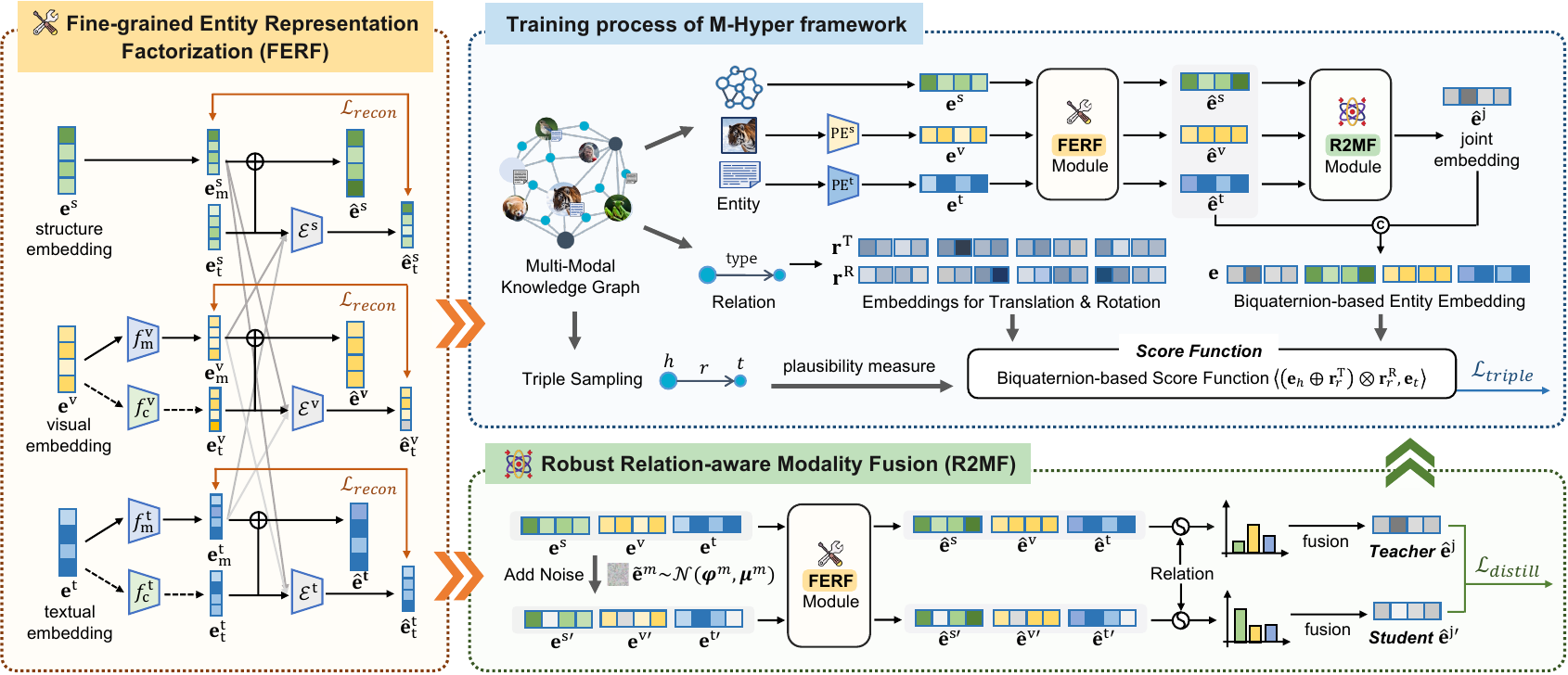}
\vspace{-1mm}
\caption{The overview of our M-Hyper, which integrates the Fine-grained Entity Representation Factorization (FERF) module and the Robust Relation-aware Modality Fusion (R2MF) module to learn robust representations for three modalities and their fusion, enabling unified multi-modal knowledge graph modeling in hypercomplex spaces.} 
\vspace{-1mm}
\label{method}
\end{figure*}

\section{Methodology}
In this section, we introduce \textbf{M-Hyper}, which models \textbf{\underline{M}}ulti-modal knowledge graphs (MMKG) in \textbf{\underline{Hyper}}complex spaces. As shown in Figure~\ref{method}, we utilize the Fine-grained Entity Representation Factorization (FERF) module and the Robust Relation-aware Modality Fusion (R2MF) module to obtain robust representations for three independent modalities and one fused modality. These modality representations are mapped to the four orthogonal bases of a biquaternion, enabling unified score modeling.

\subsection{Problem Definition}
A Multi-modal Knowledge Graph (MMKG) can be denoted as $\mathcal{G}=(\mathcal{E}, \mathcal{R}, \mathcal{T})$, where $\mathcal{E}, \mathcal{R}$ are the entity set and relation set, and $\mathcal{T}=\{(h, r, t)| h, t\in\mathcal{E}, r\in\mathcal{R}\}$ represents the set of triples. Additionally, for each entity $e\in \mathcal{E}$, its modality information can be denoted as $\mathcal{X}^m(e)$ under a specific modality $m\in\mathcal{M}$. Specifically, $\mathcal{X}^m(e)$ can be a set of image or textual description for entity $e$, or structural information embodied in the KG $\mathcal{G}$.

Multi-modal Knowledge Graph Completion (MMKGC) models measure the plausibility of each triple $(h, r, t)\in\mathcal{T}$ using a score function $\phi$ to embed the entities and relations into a continuous vector space. We usually evaluate MMKGC models with the link prediction task, which requires predicting the missing head entity or tail entity for a given query $(?, r, t)$ or $(h, r, ?)$. For each candidate $e\in\mathcal{E}$, the score of the triple $(h, r, e)$ or $(e, r, t)$ is calculated and then ranked across the entire candidate entity set.

\subsection{Fine-grained Entity Representation Factorization}
Modality missing~\cite{missing} and cross-modal semantic ambiguity~\cite{mygo} consistently challenge the robustness of MMKGC models. These issues not only lead to information inconsistency across modalities but also introduce significant noise, making it more difficult to extract task-relevant semantic information, especially in scenarios requiring modality-cooperative reasoning. To address this problem, we decompose the representation of each individual modality $m$ into two complementary semantic subspaces: (1) modality-specific representation $\mathbf{e}^m_\text{m}\in\mathbb{R}^{2d}$ and (2) task-specific representation $\mathbf{e}^m_\text{t}\in\mathbb{R}^{2d}$.

For modality-specific representation $\mathbf{e}^m_\text{m}$, the structural embedding $\mathbf{e}_\text{m}^{\text{s}}$ is learned from scratch during training, while textual and visual modality embeddings are learned from the features extracted by the pre-trained model~\cite{BERT,VGG}, denoted as:
\begin{equation}
    \mathbf{e}^m_\text{m}=f^{m}_\text{m}(\frac{1}{\left | \mathcal{X}^m(e) \right | } \sum_{x^m\in\mathcal{X}^m(e)} \text{PE}^m(x^m)),
\end{equation}
where $m\in\{\text{t},\text{v}\}, f^{m}_\text{m}:\mathbb{R} ^{d^{m}}\rightarrow \mathbb{R}^{2d}$ is 1-layer MLP, $\mathcal{X}^m(e)$ is the set of modality information for $m$ modality of entity $e$, and $\text{PE}^m$ represents the pre-trained encoder. For task-specific representation $\mathbf{e}^m_\text{t}$, they are all learnable embeddings during training. Among them, visual $\mathbf{e}^\text{v}_\text{t}$ and textual $\mathbf{e}^\text{t}_\text{t}$ embeddings are initialized by applying PCA to extract coarse-grained modal information from raw embeddings.

Furthermore, to ensure task-specific representations not only retain the unique characteristics of each independent modality but also enhance cross-modal collaborative representation capabilities, we introduce a reconstruction loss:
\begin{equation}
    \mathcal{L}_{recon}=\sum_m||\mathcal{E}^m(\mathbf{e}^m_\text{t};\{\mathbf{e}_\text{m}^{\hat{m}}:\hat{m}\neq m\})-\mathbf{e}^m_\text{m}||^2,
\end{equation}
where $\mathcal{E}^m$ is MLP. This loss requires the modality-specific embeddings to collaborate with other modalities to jointly reconstruct the original modality information. The final embedding is $\mathbf{\hat{e}}^m=\mathbf{e}^m_\text{m}+\mathbf{e}^m_\text{t}$ for modality $m$, and entire module can be denoted as: $\mathbf{\hat{e}}^{\textbf{s}},\mathbf{\hat{e}}^{\text{v}},\mathbf{\hat{e}}^{\text{t}}=\mathrm{FERF}(\mathbf{e}^{\textbf{s}},\mathbf{e}^{\text{v}},\mathbf{e}^{\text{t}})$.

\subsection{Robust Relation-aware Modality Fusion}
In terms of relation representation, to model both translation and rotation transformations~\cite{duale}, we define Translation embeddings $\textbf{r}^\text{T}=||_{i=1}^{4} \mathbf{r}_i^\text{T} \in\mathbb{R}^{8d}$ and Rotation embeddings $\textbf{r}^\text{R}=||_{i=1}^{4} \mathbf{r}_i^\text{R}\in\mathbb{R}^{8d}$ for each relation $r$, and their algebraic representations are denoted as $Q_r^\text{T}$ and $Q_r^\text{R}$.

\paragraph{Relation-aware Gated Fusion.}Considering that modality information required by an entity varies across different relation queries, we aim to design a adaptive relation-aware fusion strategy for the entity's fused modality embeddings $\mathbf{\hat{e}}^\text{j}$. Specifically, we first compute the contribution weights of the entity's modality embeddings under relation $r$:
\begin{equation}
w^{m}=f^m_w([\mathbf{\hat{e}}^{m};\mathbf{r}^\text{T};\mathbf{r}^\text{R}]), m\in\{\text{s},\text{v},\text{t}\}
\end{equation}
where $f^{m}_w: \mathbb{R}^{18d}\rightarrow \mathbb{R}^1$ are 1-layer MLPs. Then, when applying softmax to normalize the weights, we introduce a learnable relation-wise temperature coefficient $\tau_r$ to further optimize the weight distribution: $\hat{w}^{m}(e,r) = \frac{\exp({w}^{m} / \tau_r)}{\sum_i \exp({w}^{i} / \tau_r)}.$ Consequently, during gated fusion process, we also equip entity $e$ with a task-specific embedding ${\mathbf{e}}_\text{t}^{\text{j}}\in\mathbb{R}^{2d}$ denoted as:
\begin{equation}
    \mathbf{\hat{e}}^{\text{j}}=\sum_m \hat{w}^{m}\mathbf{\hat{e}}^{m}+\mathbf{e}_\text{t}^\text{j}, m\in\{\text{s},\text{v},\text{t}\}.
\end{equation}
Ultimately, we denote the entire relation-aware gated fusion process as: $\mathbf{\hat{e}}^\text{j}=\mathrm{Rel}(\mathbf{\hat{e}}^{\textbf{s}},\mathbf{\hat{e}}^{\text{v}},\mathbf{\hat{e}}^{\text{t}})$.

\paragraph{Noise-powered Self-distillation.}\citet{missing} have found introducing a certain degree of modality noise into MMKGs can effectively enhance the robustness of the model's entity representations. Inspired by this, we aim to enhance the robustness of dynamic gated fusion by introducing modality noise. Specifically, given the original embedding set $\{\mathbf{e}_i^m\}_{i=1}^N$ of modality $m$, we can calculate the feature mean $ \boldsymbol{\varphi}^m = \frac{1}{N} \sum_{i=1}^N \mathbf{e}_i^m$ and variance $\boldsymbol{\mu}^m = \frac{1}{N} \sum_{i=1}^N (\mathbf{e}_i^m-\boldsymbol{\varphi}_m)^2$. Next, we add Gaussian noise $\tilde{\mathbf{e}}^m \sim \mathcal{N}(\boldsymbol{\varphi}^m, \boldsymbol{\mu}^m)$ to a certain ratio $\beta$ of original representations, denoted as: $\mathbf{e}^{\text{s}}{}'=\mathbf{e}^{\text{s}}+\tilde{\mathbf{e}}^m$. Furthermore, we take the fused embedding obtained without noise $\mathbf{\hat{e}}_i^\text{j}$ as teacher and the fused embedding obtained with added noise $\mathbf{\hat{e}}_i^\text{j}{}'$ as the student. During the training process, a self-distillation loss is introduced: 
\begin{equation}
\mathcal{L}_{distill} = \frac{1}{n} \sum_{i=1}^{n} \left\| \mathbf{\hat{e}}_i^\text{j} - \mathbf{\hat{e}}_i^\text{j}{}' \right\|^2,
\end{equation}
where $\mathbf{\hat{e}}_i^\text{j}=\mathrm{Rel}(\mathrm{FERF}(\mathbf{{e}}^{\textbf{s}},\mathbf{{e}}^{\text{v}},\mathbf{{e}}^{\text{t}}))$ is teacher embedding and $\mathbf{\hat{e}}_i^\text{j}{}'=\mathrm{Rel}(\mathrm{FERF}(\mathbf{{e}}^{\textbf{s}}{}',\mathbf{{e}}^{\text{v}}{}',\mathbf{{e}}^{\text{t}}{}'))$ is student embedding. They share parameters of $\mathrm{Rel}$ and $\mathrm{FERF}$ modules. Noise-powered perturbations enforce embedding consistency and enhance the fusion gate's noise robustness. 

\begin{figure*}
\centering
\includegraphics[scale=0.67]{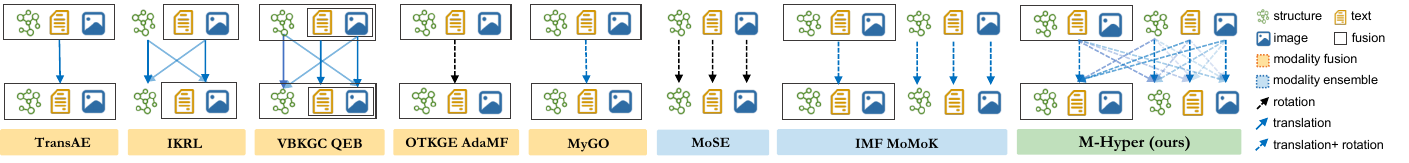}
\vspace{-5mm}
\caption{Compared to existing MMKGC score functions, M-Hyper achieves the most comprehensive modality interaction and geometric transformation. For detailed formulaic theoretical proofs, please refer to Appendix A.} \label{interaction}
\vspace{-1mm}
\end{figure*}

\subsection{Training with Biquaternion-based Score Function}

To enable the coexistence of one fused and three independent modalities, we represent them respectively as the real part and the three imaginary parts of a biquaternion for entity $e$. Its \textit{\textbf{algebraic representation}} is: $Q=\hat{\textbf{e}}^{\text{j}} + \hat{\textbf{e}}^{\text{s}}\mathbf{i} + \hat{\textbf{e}}^{\text{v}}\mathbf{j} + \hat{\textbf{e}}^{\text{t}}\mathbf{k} =(\hat{\textbf{e}}_{\text{r}}^{\text{j}}+\hat{\textbf{e}}_{\text{i}}^{\text{j}}\mathbf{I}) + (\hat{\textbf{e}}_{\text{r}}^{\text{s}}+\hat{\textbf{e}}_{\text{i}}^{\text{s}}\mathbf{I})\mathbf{i} +(\hat{\textbf{e}}_{\text{r}}^{\text{v}}+\hat{\textbf{e}}_{\text{i}}^{\text{v}}\mathbf{I})\mathbf{j} + (\hat{\textbf{e}}_{\text{r}}^{\text{t}}+\hat{\textbf{e}}_{\text{i}}^{\text{t}}\mathbf{I})\mathbf{k}$, whose all coefficients $\hat{\mathbf{e}}^m=[\hat{\mathbf{e}}^m_\text{r};\hat{\mathbf{e}}^m_\text{i}]\in\mathbb{R}^{2d}$ are parameterized as embeddings, and \textit{\textbf{embedding representation}} is the concatenation of all coefficients, denoted as:
\begin{equation}
\begin{split}
    \textbf{e} &= [\textbf{e}^{\text{j}};\textbf{e}^{\text{s}};\textbf{e}^{\text{v}};\textbf{e}^{\text{t}}] \\
    & =[\textbf{e}_{\text{r}}^{\text{j}};\textbf{e}_{\text{i}}^{\text{j}};\textbf{e}_{\text{r}}^{\text{s}};\textbf{e}_{\text{i}}^{\text{s}};\textbf{e}_{\text{r}}^{\text{v}};\textbf{e}_{\text{i}}^{\text{v}};\textbf{e}_{\text{r}}^{\text{t}};\textbf{e}_{\text{i}}^{\text{t}}]\in\mathbb{R}^{8d}
\end{split}
\end{equation} 
Considering that the imaginary units of the quaternion field $\mathbb{H}$ are symmetric, the permutation of these modalities does not affect the representation. 

\paragraph{Biquaternion-based Score Function.}We adopt a standard semantic-matching~\cite{pami,bique} strategy to score the plausibility of triple. For a given triple $(h, r, t)$, we first apply the following algebraic operation to calculate the embedding of query $(h,r,?)$:
\begin{equation}
\label{11}
Q_{h''} = ( Q_h \oplus Q_r^\text{T} ) \otimes Q_r^\text{R},
\end{equation}
where $\oplus$ and $\otimes$ represent addition and Hamilton product between biquaternions. The addition is an element-wise sum: $Q_{h'} = Q_h \oplus Q_r^\text{T}=(\textbf{e}_{h}^{\text{j}}+\textbf{r}_1^{\text{T}}) + (\textbf{e}_{h}^{\text{s}}+\textbf{r}_2^{\text{T}})\mathbf{i} + (\textbf{e}_{h}^{\text{v}}+\textbf{r}_3^{\text{T}})\mathbf{j} + (\textbf{e}_{h}^{\text{t}}+\textbf{r}_4^{\text{T}})\mathbf{k} = \textbf{e}_{h'}^{\text{j}} + \textbf{e}_{h'}^{\text{s}}\mathbf{i} + \textbf{e}_{h'}^{\text{v}}\mathbf{j} + \textbf{e}_{h'}^{\text{t}}\mathbf{k}$ for characterizing translation transformations. Then $Q^{\text{T}}_r$ rotates the query via Hamilton product as shown in Equation~\ref{product}:
\begin{equation}
Q_{h''} = Q_{h'} \otimes Q_r^\text{R} = \sum_{m}^{|\mathcal{M}|} \sum_{k=1}^{4} H_{ikm} (\mathbf{e}_{h'}^m) \circledast(\mathbf{r}^\text{R}_k) \, \mathbf{u}_i,
\end{equation}
where $H_{ikm}$ denotes the structure constants uniquely defining the multiplication rules for biquaternion algebra, and $\mathbf{u}_i \in\{1,\mathbf{i},\mathbf{j},\mathbf{k}\}$ indicates the corresponding basis element, and $\circledast$ represents multiplication in complex number field $\mathbb{C}$. 

\paragraph{Optimization Objective.}We utilize the standard vector dot-product between query $Q^{\prime\prime}_h$ and tail entity $Q_t=\textbf{e}_{t}^{\text{j}} + \textbf{e}_{t}^{\text{s}}\mathbf{i} + \textbf{e}_{t}^{\text{v}}\mathbf{j} + \textbf{e}_{t}^{\text{t}}\mathbf{k}$ to compute plausibility score: $\phi(h, r, t)=\langle Q_{h''},Q_t\rangle =[\textbf{e}_{h''}^\text{j};\textbf{e}_{h''}^{\text{s}};\textbf{e}_{h''}^{\text{v}};\textbf{e}_{h''}^{\text{t}}] \cdot [\textbf{e}_{t}^{\text{j}};\textbf{e}_{t}^{\text{s}};\textbf{e}_{t}^{\text{v}};\textbf{e}_{t}^{\text{t}}]^{\top}$. We optimize our model using the cross-entropy loss:
\begin{equation}
    \mathcal{L}_{triple} =\sum_{t^\prime}^{| \mathcal{V}|}\log(1+\text{exp}(y_{t^\prime}\phi(h,r,t^\prime)),
\end{equation}
where $y_{t^\prime}$ is the ground-truth of the candidate tail entity $t^\prime$. So the Biquaternion-based score function can be expressed as: $\phi(h, r, t)= \langle (Q_h \oplus Q_r^\text{T} ) \otimes Q_r^\text{R},Q_t\rangle$. As shown in Figure~\ref{interaction}, it can achieve the most comprehensive modal interaction and geometric transformation (translation + rotation). We provide a more in-depth theoretical proof in Appendix A.
The overall training objective $\mathcal{L}_{total}$ is represented as:
\begin{equation}
\mathcal{L}_{total}=\mathcal{L}_{recon}+\mathcal{L}_{distill}+\mathcal{L}_{triple}+\mathcal{L}_{reg},
\end{equation}
where we also employ N3 regularization norm~\cite{N3} to prevent overfitting: $\mathcal{L}_{reg}=\lambda(||\mathbf{e}_h||_3^3+||\mathbf{r}_r^\text{T}||_3^3+||\mathbf{r}_r^\text{R}||_3^3+||\mathbf{e}_t||_3^3)$
, and $\lambda$ is a regularization hyperparameter.

\begin{table*}[ht]
\centering
\resizebox{0.98\textwidth}{!}{
\begin{tabular}{cl|cccc|cccc|cccc}
\toprule
\multicolumn{2}{c|}{\multirow{2}{*}{Model}} & \multicolumn{4}{c|}{DB15K} & \multicolumn{4}{c|}{MKG-W} & \multicolumn{4}{c}{MKG-Y} \\
\multicolumn{2}{c|}{} & MRR & Hit@1 & Hit@3 & Hit@10& MRR & Hit@1 & Hit@3 & Hit@10& MRR & Hit@1 & Hit@3 & Hit@10\\
\midrule
\multicolumn{1}{c}{\multirow{5}{*}{\begin{tabular}[c]{@{}c@{}}Uni-modal\\ KGC\end{tabular}}}& TransE & 24.86 & 12.78 & 31.48 & 47.07 & 29.19 & 21.06 & 33.20& 44.23 & 30.73 & 23.45 & 35.18 & 43.37 \\
\multicolumn{1}{c}{} & ComplEx  & 27.48 & 18.37 & 31.57 & 45.37 & 24.93 & 19.09 & 26.69 & 36.73 & 28.71 & 22.26 & 32.12 & 40.93 \\
\multicolumn{1}{c}{} & RotatE & 29.28 & 17.87 & 36.12 & 49.66 & 33.67 & 26.80 & 36.68 & 46.73 & 34.95 & 29.10 & 38.35 & 45.30 \\
\multicolumn{1}{c}{} & QuatE$^{*}$ & 34.18 & 25.42 & 38.91 & 51.30 & 34.50& 28.94 & 36.71 & 46.64 & 36.01 & 30.53 & 38.84 & 43.68 \\
\multicolumn{1}{c}{} & DualE$^{*}$  & 35.85 & 29.31 & 38.52 & 51.28 & 33.94 & 27.55 & 36.56 & 46.09 & 34.95 & 29.77 & 38.44 & 43.12 \\
\multicolumn{1}{c}{} & BiQUE$^{*}$  & 38.34 & 32.38 & 41.48 & 53.23 & 35.01 & 29.42 & 37.01 & 46.49 & 36.74 & 34.82 & 38.25 & 42.16 \\
\midrule
\multicolumn{1}{c}{\multirow{13}{*}{\begin{tabular}[c]{@{}c@{}}Multi-modal\\ KGC\end{tabular}}}& IKRL & 26.82 & 14.09 & 34.93 & 49.09 & 32.36 & 26.11 & 34.75 & 44.07 & 33.22 & 30.37 & 34.28 & 38.26 \\
\multicolumn{1}{c}{} & TransAE & 28.09 & 21.25 & 31.17 & 41.17 & 30.00 & 21.23 & 34.91 & 44.72 & 28.10 & 25.31 & 29.10 & 33.03 \\
\multicolumn{1}{c}{} & VBKGC & 30.61 & 19.75 & 37.18 & 49.44 & 30.61 & 24.91 & 33.01 & 40.88 & 37.04 & 33.76 & 38.75 & 42.30 \\
\multicolumn{1}{c}{} & OTKGE & 23.86 & 18.45 & 25.89 & 34.23 & 34.36 & 28.85 & 36.25 & 44.88 & 35.51 & 31.97 & 37.18 & 41.38 \\
\multicolumn{1}{c}{} & MoSE & 28.38 & 21.56 & 30.91 & 41.67 & 33.34 & 27.78 & 33.94 & 41.06 & 36.28 & 33.64 & 37.47 & 40.81  \\
\multicolumn{1}{c}{} & MMRNS  & 32.68 & 23.01 & 37.86 & 51.01 & 35.03 & 28.59 &37.49 &47.47 & 35.93 & 30.53 & 39.07 & \underline{45.47} \\
\multicolumn{1}{c}{} & QEB & 28.18 & 14.82 & 36.67 & 51.55 & 32.38 & 25.47 & 35.06 & 45.32 & 34.37 & 29.49 & 36.95 & 42.32 \\
\multicolumn{1}{c}{} & VISTA & 30.42 & 22.49 & 33.56 & 45.94 & 32.91 & 26.12 & 35.38 & 45.61 & 30.45 & 24.87 & 32.39 & 41.53 \\
\multicolumn{1}{c}{} & IMF  & 32.25 & 24.20 & 36.00 & 48.19 & 34.50 & 28.77 & 36.62 & 45.44 & 35.79 & 32.95 & 37.14 & 40.63  \\
\multicolumn{1}{c}{} & AdaMF & 32.51 & 21.31 & 39.67 & 51.68 & 34.27 & 	27.21	& 37.86 & 47.21 & 38.06 & 33.49 & \underline{40.44} & \textbf{45.48}\\
\multicolumn{1}{c}{} & MyGO & 37.72 & 30.08 & 41.26 & 52.21 & \underline{36.10} & 	29.78	& \underline{38.54} & \underline{47.75} & \underline{38.44} & 35.01 & 39.84 & 44.19\\
\multicolumn{1}{c}{} & K-ON$^{*}$ & 36.24 &28.13	&40.49	&51.26	&35.83	&29.41	&37.32	&47.16	&35.83	&32.56	&37.34	&42.45\\
\multicolumn{1}{c}{} & MoMoK & \underline{39.57} & \underline{32.38} & \underline{43.45} & \underline{54.14} & 35.89 & \underline{30.38}	& 37.54 & 46.13 & 37.91 & \underline{35.09} & 39.20 & 43.20\\


\midrule
\multicolumn{1}{c}{\multirow{1}{*}{\textbf{Ours}}}& \textbf{M-Hyper}&\textbf{41.25}	 &\textbf{33.64}	&\textbf{45.01}	&\textbf{56.09}	&\textbf{37.02}	&\textbf{31.24}	&\textbf{39.16}	&\textbf{48.84}	&\textbf{39.46}	&\textbf{36.02}	&\textbf{40.92}	&45.22 \\
\bottomrule
\end{tabular}}
\vspace{-1mm}
\caption{Results on DB15K, MKG-W, and MKG-Y datasets. The best results are marked \textbf{bold} and the second-best results are \underline{underlined}. The $^{*}$results are reproduced by us, and others are taken from MoMoK~\cite{momok}.}
\vspace{-1mm}
\label{table:main}
\end{table*}

\section{Experiments}

\subsection{Experimental Settings}
\paragraph{Datasets.}The experiments are conducted on three common MMKG benchmarks: DB15K~\cite{MMKG}, MKG-W~\cite{MMRNS} and MKG-Y~\cite{MMRNS}. To ensure fairness in comparison with previous works, we adopt the same representations of the visual and textual modalities in the original datasets derived from the pre-trained models VGG~\cite{VGG} and BERT~\cite{BERT}. DB15K~\cite{MMKG} is a subset of DBPedia~\cite{DBPedia} with images crawled from search engines. MKG-W and MKG-Y~\cite{MMRNS} are derived from Wikidata~\cite{wikidata} and YAGO~\cite{yago} respectively. The detailed statistics are shown in Appendix F.

\paragraph{Evaluation Protocols.}Link prediction tasks need to predict the missing entity of a given query $(h,r,?)$ or $(?,r,t)$ from $\mathcal{T}_{test}$. Consistent with the existing works, We use Mean Reciprocal Rank (MRR) and Hit@K (K=1, 3, 10) to evaluate the results. MRR and Hit@K metrics can be calculated as: $\mathbf{MRR}=\frac{1}{|\mathcal{T}_{test}|}\sum_{i=1}^{|\mathcal{T}_{test}|}(\frac{1}{r_{h,i}}+\frac{1}{r_{t,i}})$, $\mathbf{Hit@K}=\frac{1}{|\mathcal{T}_{test}|}\sum_{i=1}^{|\mathcal{T}_{test}|}(\mathbf{1}(r_{h,i} \leq K)+\mathbf{1}(r_{t,i} \leq K))$, where $r_{h,i}$ and $r_{t,i}$ are the results of head prediction and tail prediction respectively. Besides, we apply filter setting~\cite{TransE} to eliminate existing facts in the dataset. 

\paragraph{Baselines.}We select 19 representative MMKGC methods as our baselines, including: \textbf{(1) Uni-modal KGC methods}: TransE~\cite{TransE},  ComplEx~\cite{complEx}, RotatE~\cite{RotatE}, QuatE~\cite{quate}, DualE~\cite{duale} ,and BiQUE~\cite{bique}. These methods only model structural information of the KGs. \textbf{(2) Multi-modal KGC models}: \textit{fusion-based} methods: IKRL~\cite{IKRL}, TransAE~\cite{TransAE}, VBKGC~\cite{VBKGC}, OTKGE~\cite{OTKGE}, QEB~\cite{QEB}, VISTA~\cite{VISTA}, AdaMF~\cite{AdaMF-MAT}, MyGO~\cite{mygo}, K-ON~\cite{k-on}, \textit{ensemble-based} methods: MoSE~\cite{MOSE}, IMF~\cite{IMF}, MoMoK~\cite{momok}. These methods utilize both the structural information and multi-modal information in the KGs, among which K-ON~\cite{k-on} is the most advanced LLM-based method.

\paragraph{Implementation Details.}
All experiments are conducted on a Nvidia A800 GPU and implemented with PyTorch. We also add inverse triple $(t, r^{-1}, h)$ for each observed triple $(h, r, t)$ in trainset as training samples. We use Adagrad~\cite{adagrad} as the optimizer. For hyperparameters, batch size is fixed at 1000; and we search the learning rate $\alpha\in\{\mathbf{0.1},0.05,0.01,0.005\}$; dimension of embeddings $d\in\{64,\mathbf{128},256\}$; regularization factors $\lambda\in\{0.01,\mathbf{0.005},0.001\}$ and noise rate $\beta\in\{0.1,\mathbf{0.2},0.4\}$. 

\subsection{Main Results}
The experimental results are shown in Table \ref{table:main}. M-Hyper outperforms 18 existing baselines on most metrics, including AdaMF and MoMoK, which also adopt modality noise enhancement. Specifically, M-Hyper achieves a 4.25\% improvement in MRR and a 3.89\% improvement in Hit@10, demonstrating significant performance improvements. Compared to the classic fusion-based~\cite{IMF} and ensemble-based~\cite{mygo} paradigm, M-Hyper not only preserves the original modality information but also enables dynamic and flexible modality interaction, providing a promising modeling paradigm for MMKGC task.

\begin{figure}[ht]
\centering
\includegraphics[scale=0.57]{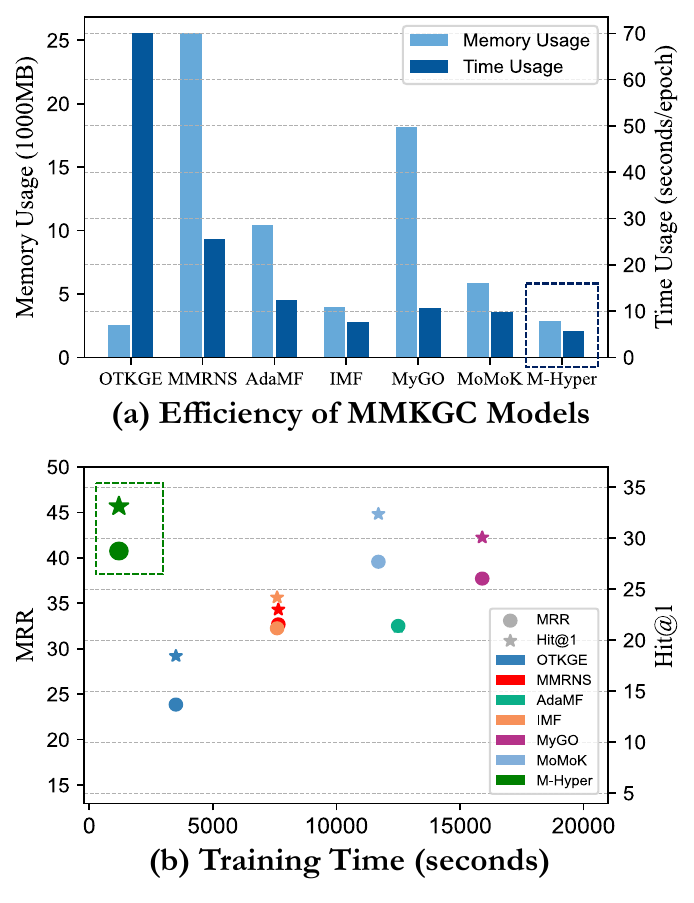}
\vspace{-3mm}
\caption{Efficiency results on memory usage, training time usage, and the trade-off between training effectiveness and training time on DB15K dataset.} \label{efficiency}
\vspace{-2mm}
\end{figure}

\subsection{Efficiency Analysis}
We conduct an efficiency analysis of M-Hyper focusing on memory usage and runtime, with the results shown in Figure~\ref{efficiency}(a). Compared to 6 state-of-the-art methods, our M-Hyper achieves the best training efficiency, requiring the shortest runtime for a single training epoch. In terms of memory usage, M-Hyper demonstrates nearly optimal performance. Figure~\ref{efficiency}(b) illustrates the training times required to achieve the best performances. Our model requires only 1160 seconds of training time to achieve an MRR of 40.75\% and a Hit@1 of 33.14\%. So we can conclude M-Hyper not only delivers the best performance but also achieves the highest computational efficiency with the least memory usage and the shortest training time.

\subsection{Ablation Study}
\begin{table}
\resizebox{0.48\textwidth}{!}{
\begin{tabular}{cl|cccccc}
\toprule
\multicolumn{2}{c|}{\multirow{2}{*}{Setting}}                         & \multicolumn{2}{c}{DB15K} & \multicolumn{2}{c}{MKG-W} & \multicolumn{2}{c}{MKG-Y} \\ 
\multicolumn{2}{c|}{}                                                  & MRR         & Hit@1       & MRR         & Hit@1       & MRR         & Hit@1       \\ \midrule
\multirow{4}{*}{$\mathcal{G}_0$} & w/o joint $\mathbf{\hat{e}}^{\text{j}}$       & 36.36       & 28.54       & 35.09       & 29.16       & \underline{36.71}       & \underline{33.42}       \\
  & w/o structure $\mathbf{\hat{e}}^{\text{s}}$   & 39.77       & 32.17       & \underline{34.62}       & \underline{28.63}       & 38.03       & 34.60       \\
    & w/o vision $\mathbf{\hat{e}}^{\text{v}}$      & \underline{35.09}       & \underline{27.22}       & 36.46       & 30.60       & 37.95       & 34.68       \\
    & w/o text $\mathbf{\hat{e}}^{\text{t}}$        & 39.70       & 32.12       & 36.28       & 31.17       & 38.09       & 34.74       \\ \midrule
\multirow{5}{*}{$\mathcal{G}_1$}   & w/o FERF   &39.24       & 31.83       & 35.93      & 29.38    & 37.93       & 34.53  \\
& w/o noise-powered        & 39.64       & 32.16       & 36.10     & 30.28    & 38.16       & 35.82        \\
& w/o $r$-aware gate       & 40.18       & 32.47       & 36.18       & 30.44       & 38.21       & 35.14       \\
& w/o $\mathcal{L}_{recon}$       & 40.97       & 33.24       & 36.18       & 30.69       & 39.12       & 35.23       \\
\multirow{2}{*}{}    & w/o translation $\mathbf{r}^\text{T}$ & 39.50       & 31.42       & 35.13       & 29.56       & 37.86       & 34.64       \\
& w/o rotation $\mathbf{r}^\text{R}$    & 38.91       & 31.35       & 36.46       & 30.67       & 37.78       & 34.55       \\
& M-Hyper+DualE    & 39.93       & 32.07       & 35.96       & 30.10       & 38.02       & 34.78       \\ \midrule
& {M-Hyper-fusion}            & 39.23 &31.66  &35.54 &30.35 &37.52 &34.51      \\
$\mathcal{G}$                         & {M-Hyper-ensemble}            & 39.31 &31.71  &34.75 &29.26 &37.58 &34.78      \\
& \textbf{M-Hyper}(ours)            & \textbf{41.25}       & \textbf{33.64}       & \textbf{37.02}       & \textbf{31.24}       & \textbf{39.46}       & \textbf{36.02}      \\ \bottomrule
\end{tabular}}
\vspace{-1mm}
\caption{Results of modality ablation $\mathcal{G}_0$ and model ablation $\mathcal{G}_1$. $\mathcal{G}$ represents the comparison among the three modality modeling paradigms.}
\vspace{-1mm}
\label{table:ablation}
\end{table}

\paragraph{Modality Ablation Study.}To verify the contributions of each modality, we set the corresponding modality embedding to an all-zero embedding, removing its influence. As shown in Table~\ref{table:ablation}, all modalities positively impact performance, albeit to varying degrees across different datasets. Notably, excluding the joint modality leads to the most substantial performance decline, highlighting its pivotal role in M-Hyper’s overall effectiveness.

\paragraph{Model Ablation Study.}We can see that each module contributes to the overall performance. FERF and noise-powered self-distillation modules enable more robust modality representations, while the relation-aware gate facilitates dynamic modality fusion to handle complex contexts. Additionally, translation and rotation relation embeddings enable more sophisticated relational modeling. Notably, removing the rotation operation $\mathbf{r}^\text{R}$ in complex field $\mathbb{C}$ reduces hypercomplex space to a quaternion space and results in a performance decline, indicating that the biquaternion space offers greater expressive power. Meanwhile, we introduce M-Hyper variants under the ensemble and fusion paradigms, whose score functions are provided in Appendix C. It can be observed that M-Hyper, benefiting from adequate collaboration between independent and fused modalities, achieves the best performance.

\begin{figure}
\flushleft
\includegraphics[scale=0.6]{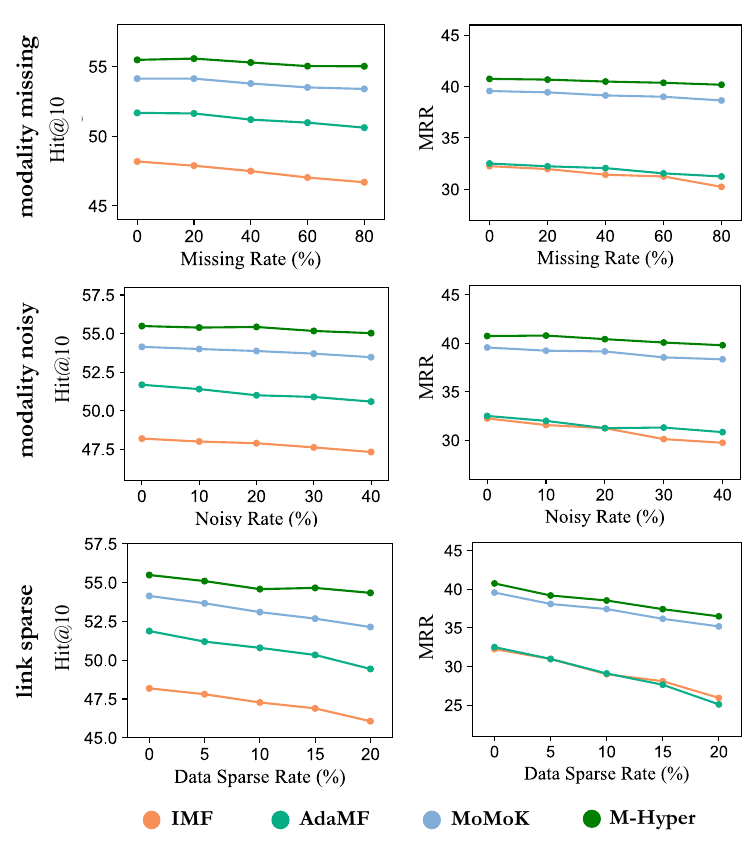}
\vspace{-5mm}
\caption{Results on DB15K under 3 complex scenarios: modality missing, modality noisy and link sparse.} 
\vspace{-3mm}
\label{complex}
\end{figure}

\subsection{Robustness to Complex Scenarios}
Following~\citet{momok}, we evaluate MMKGC robustness under three challenging scenarios: (1) modality missing, (2) modality noise, and (3) link sparsity. To be specific, in modality missing scenario, we randomly delete a certain ratio of entity's raw modality embeddings. For the modality noise scenario, we randomly add Gaussian noise to raw modality embeddings. In the link sparsity scenario, we randomly remove a certain ratio of training triples. 

\begin{figure*}[ht]
\centering
\includegraphics[scale=0.55]{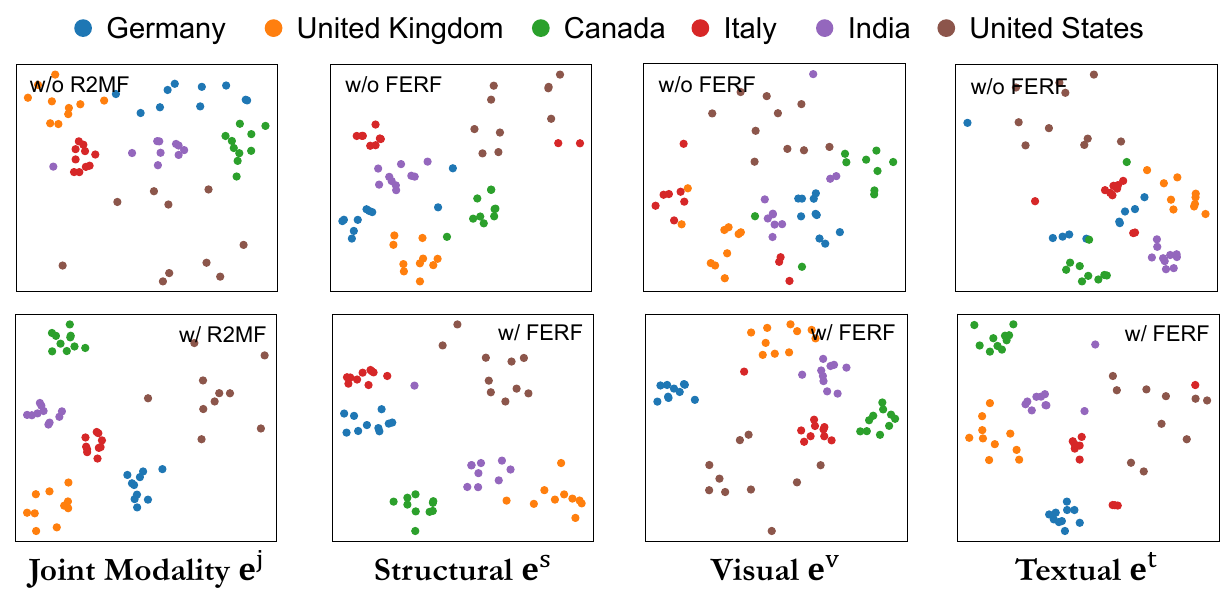}
\vspace{-3mm}
\caption{Embedding visualization under t-SNE for cities under relation \textit{country}, and distinct colors are utilized to represent different countries.}
\vspace{-2mm}
\label{visualize}
\end{figure*}

As shown in Figure~\ref{complex}, the model's performance declines to varying degrees under these complex scenarios. Among them, the training data, as a critical source of structural information, significantly contributes to the model's performance. Notably, we find that AdaMF, MoMoK, and M-Hyper with noise-augmented training achieve improved robustness. Moreover, unlike previous noise-augmented methods, we introduce task-specific representations and a self-distillation supervision strategy, which further enhance model's noise-reduction capabilities and improve the effectiveness of dynamic fusion. As a result, our approach achieves relatively superior robust performance.  

\subsection{Modality Visualization Analysis}
As illustrated in Figure~\ref{visualize}, we apply t-SNE to visualize the modality embeddings of cities across 6 countries in DB15K dataset. It is evident that the presence of modality ambiguity and bias introduces variability in the representation efficacy of entities across different modalities. Notably, among all modalities, the joint modality representations demonstrate the highest discriminative capability in differentiating entities. Furthermore, the integration of the FERF and R2MF modules significantly improves the expressiveness and effectiveness of the modality-specific embeddings, highlighting their ability to mitigate modality bias and enhance representation quality. 


\section{Conclusion}
In this paper, we highlight the limitations of existing MMKGC paradigms, which struggle to balance fused and independent modality representations. To enable efficient and flexible cross-modal collaboration, we propose M-Hyper, the first method to represent MMKGs in hypercomplex space. Specifically, we introduce Fine-grained Entity Representation Factorization (FERF) module and Robust Relation-aware Modality Fusion (R2MF) module to obtain robust representations for three independent modalities and one fused modality. Subsequently, these modality representations are mapped onto the four orthogonal bases of a biquaternion, enabling efficient modeling of pairwise interactions and comprehensive cross-modal integration. Empirical results show that our M-Hyper demonstrate greater performance and robustness. 

\section*{Limitations}
We focus on ``transductive'' multi-modal knowledge graph completion (MMKGC) under a static setting, assuming that entities, relations, and modality information remain fixed during both training and inference. Therefore, for dynamic scenarios with entities, relations, or modality features (e.g., newly added images or textual descriptions) undergoing frequent updates, it may be necessary to design online learning frameworks or dynamic modeling approaches to address evolving data distributions and incremental modality adaptation. In addition, we also hope to explore the idea of coexistence of independence and integration in other task scenarios, such as entity alignment~\cite{WOS:001551659500001}, named entity recognition~\cite{WOS:001446505200005}, and knowledge graph question answering~\cite{temp-r1}.

\section*{Ethics Statement}
In this paper, we explore the multi-modal knowledge graph completion task with deep learning techniques. Our training and evaluation are based on publicly available and widely used datasets of different types of knowledge graphs. Therefore, we believe this does not violate any ethics. 

\section*{Acknowledgments}
This work is founded by National Natural Science Foundation of China (NSFCU23B2055/NSFC62306276), New Generation Artificial Intelligence-National Science and Technology Major Project 2030 (2025ZD0122800), Yongjiang Talent Introduction Programme (2022A-238-G), and Fundamental Research Funds for the Central Universities (226-2023-00138). This work was supported by Ant Group. 

\bibliography{acl}

\appendix

\section*{Appendix}
\input{appendix}

\end{document}

%% file: appendix.tex
\subsection*{A Detailed Proof of Theorem}

\paragraph{Theorem 1:}
Let $X = \{M_\text{s}, M_\text{v}, M_\text{t}\}$ represent the multi-modal input and $Y$ the target task. The M-Hyper representation is defined as: $Q = T_{\text{j}} \mathbf{1} + T_\text{s} \mathbf{i} + T_\text{v} \mathbf{j} + T_\text{t} \mathbf{k},$
where $T_\text{j}$ encodes fused information across modalities, and $T_\text{s}, T_\text{v}, T_\text{t}$ preserve modality-specific information. Under the Information Bottleneck (IB)~\cite{ib} framework, with the IB loss:
\[
\mathcal{L}_{\text{IB}}(T) = I(X; T) - \beta I(T; Y),
\]
the M-Hyper representation achieves a strictly lower IB loss:
\begin{equation}
    \mathcal{L}_{\text{IB}}(Q) < \min\left(\mathcal{L}_{\text{IB}}(T_f),\ \mathcal{L}_{\text{IB}}(T_{\text{ens}})\right),
\end{equation}
where $T_f$ is the fused representation and $T_{\text{ens}}$ the ensemble representation.

\paragraph{Proof 1:}

Consider three representations: (1) M-Hyper $Q$, (2) fusion-based $T_f = f(X)$, and (3) ensemble $T_{\text{ens}} = \{T_\text{j}, T_\text{s}, T_\text{v}, T_\text{t}\}$. On the one hand, fusion $T_f$ over-compresses and includes redundancy:
\begin{align}
    I(X; T_f) - I(X; Q) &= \Delta_{\text{redundancy}} 
    \\ &= \sum_{i\ne j}I(T_i;T_j|Y) > 0,
\end{align}

where $\Delta_{\text{redundancy}}$ measures cross-modal redundancy that does not contribute to $Y$.
On the other hand, ensemble $T_{\text{ens}}$ lacks explicit interactions:
\begin{align}
    I&(T_{\text{ens}}; Y) \leq \\
    \sum_{i} I(T_i; Y)& + I(T_\text{fuse}; Y) - \sum_{i<j}I(T_i;T_j;Y),
\end{align}
where triple mutual information $\sum_{i<j}I(T_i;T_j;Y)$ captures cross-modal synergy not fully utilized in a simple ensemble. Our $Q$ in quaternion space $\mathbb{H}$ (via Hamilton product, see Theorem 1) generates interaction terms $C_{ij}=T_i \cdot T_j$ that satisfy:
\[
\sum_{i<j} I(C_{ij}; Y) \geq \eta \left\|T_i^\top T_j\right\|^2 > 0,
\]
i.e., these interactions are informative for predicting $Y$. Imposing orthogonality ($\langle T_i, T_j \rangle = 0,\, i \ne j$) further reduces intra-representation redundancy, so
\[
I(X; Q) < I(X; T_{\text{ens}}).
\]
As $Q$ contains all modality-specific information (from $T_\text{j}, T_\text{s}, T_\text{v}, T_\text{t}$) plus explicit cross-modal interactions (i.e., $C_{ij}$), it is at least as informative about $Y$ as $T_{\text{ens}}$, and typically more so:
\[
I(Q; Y) \geq I(T_{\text{ens}}; Y).
\]
Combining the above, the difference in IB loss between $Q$ and $T_{\text{ens}}$ becomes
\begin{equation}
\begin{aligned}
&\mathcal{L}_{\text{IB}}(Q) - \mathcal{L}_{\text{IB}}(T_{\text{ens}})   \\
&= \big[I(X; Q) - I(X; T_{\text{ens}})\big] \\&\quad- \beta\big[I(Q; Y) - I(T_{\text{ens}}; Y)\big]< 0 .
\end{aligned}
\end{equation}
The first term is negative (due to reduced redundancy), and the second term is non-positive (due to improved relevance); thus their sum is strictly negative under $\beta > 0$.
The comparison with $T_f$ is similar, as detailed before:
\[
\mathcal{L}_{\text{IB}}(Q) - \mathcal{L}_{\text{IB}}(T_f) \leq -\Delta_{\text{redundancy}} - \beta \Delta_{\text{interaction}} < 0,
\]
where $\Delta_{\text{interaction}} = I(Q; Y) - I(T_f; Y) \ge 0$. So we can conclude:
\begin{equation}
\mathcal{L}_{\text{IB}}(Q) < \min \big(\mathcal{L}_{\text{IB}}(T_f),\ \mathcal{L}_{\text{IB}}(T_{\text{ens}})\big)
\end{equation}
Therefore, $Q$ achieves a strictly lower IB loss by both reducing redundancy (better compression of $X$) and boosting task relevance (enhanced dependence on $Y$) via explicit cross-modal interactions.

\begin{figure*}
\centering
\includegraphics[scale=0.5]{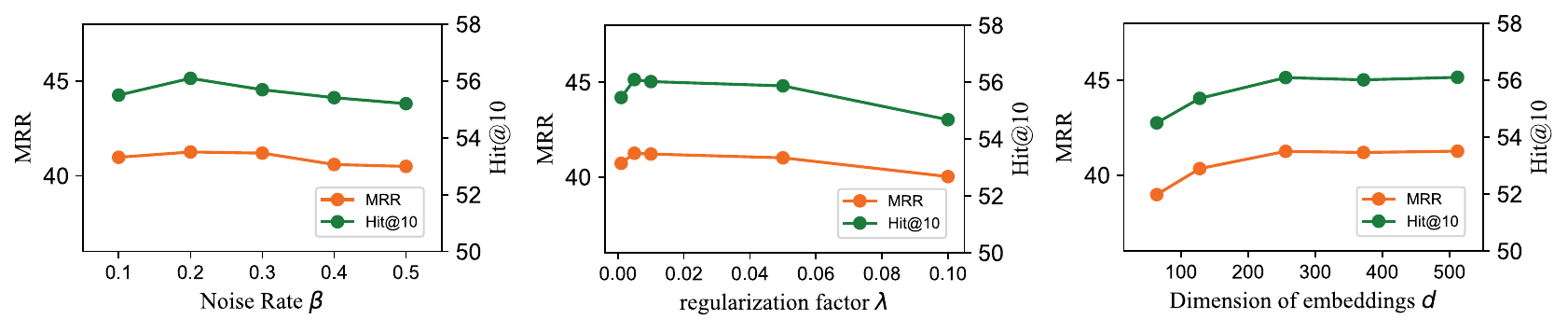}
\vspace{-2mm}
\caption{Results of hyperparamter analysis for noise rate $\beta$, regularization factor $\lambda$ and dimension $d$.} \label{hyperparameter}
\end{figure*}

\paragraph{Theorem 2:}Let the entity embedding be $Q = \textbf{e}^\text{j} \mathbf{1} + \textbf{e}^\text{s} \mathbf{i} + \textbf{e}^\text{j} \mathbf{j} + \textbf{e}^\text{t} \mathbf{k}$, and score function as:
\begin{equation}
    \phi(h, r, t)= \langle (Q_h \oplus Q_r^\text{T} ) \otimes Q_r^\text{R},Q_t\rangle
\end{equation}
Then for any modalities $ m,m'\in \{\text{j, s, v, t}\}$, the algebraic expansion contains all pair-wise interaction $\mathbf{e}_h^{m} \cdot \textbf{e}_t^{m'}$, as well as intra-modal terms.

\paragraph{Proof 2:}For the sake of simplicity in representation, we mark the final representations of each modality $m$ as: $\mathbf{e}^m$. Our biquaternion-based score function can be expanded as:
\begin{equation}
\resizebox{0.95\columnwidth}{!}{
$\begin{aligned}
& \phi(h, r, t)= \langle (Q_h \oplus Q_r^\text{T} ) \otimes Q_r^\text{R},Q_t\rangle \\
& = \langle [(\mathbf{e}_h^\text{j} + \mathbf{r}_{r,1}^\text{T})+( \mathbf{e}_h^\text{s} + \mathbf{r}_{r,2}^\text{T})\mathbf{i}+( \mathbf{e}_h^\text{v} + \mathbf{r}_{r,3}^\text{T})\mathbf{j}+( \mathbf{e}_h^\text{t} + \mathbf{r}_{r,4}^\text{T})\mathbf{k}]\\  
&\quad\otimes [\mathbf{r}_{r,1}^\text{R}+\mathbf{r}_{r,2}^\text{R}\mathbf{i}+\mathbf{r}_{r,3}^\text{R}\mathbf{j}+\mathbf{r}_{r,4}^\text{R}\mathbf{k}],[\mathbf{e}_t^\text{j}+\mathbf{e}_t^\text{s}\mathbf{i}+\mathbf{e}_t^\text{v}\mathbf{j}+\mathbf{e}_t^\text{t}\mathbf{k}]\rangle \\
& =\langle[(\mathbf{e}_h^{\text{j}}+\mathbf{r}_{r,1}^{\text{T}})\circledast \mathbf{e}_{r,1}^\text{R}-(\mathbf{e}_h^{\text{s}}+\mathbf{r}_{r,2}^{\text{T}})\circledast \mathbf{e}_{r,2}^\text{R}- \\
&\quad\; (\mathbf{e}_h^{\text{t}}+\mathbf{r}_{r,3}^{\text{T}})\circledast \mathbf{e}_{r,3}^\text{R}-(\mathbf{e}_h^{\text{v}}+\mathbf{r}_{r,4}^{\text{T}})\circledast \mathbf{e}_{r,4}^\text{R}],  \mathbf{e}_t^{\text{j}}\rangle \\
&\, +\langle[(\mathbf{e}_h^{\text{j}}+\mathbf{r}_{r,1}^{\text{T}})\circledast \mathbf{e}_{r,2}^\text{R}+(\mathbf{e}_h^{\text{s}}+\mathbf{r}_{r,2}^{\text{T}})\circledast \mathbf{e}_{r,3}^\text{R}+ \\
&\quad\; (\mathbf{e}_h^{\text{t}}+\mathbf{r}_{r,3}^{\text{T}})\circledast \mathbf{e}_{r,4}^\text{R}-(\mathbf{e}_h^{\text{v}}+\mathbf{r}_{r,4}^{\text{T}})\circledast \mathbf{e}_{r,1}^\text{R}], \mathbf{e}_t^{\text{s}}\rangle  \\
&\, +\langle[(\mathbf{e}_h^{\text{j}}+\mathbf{r}_{r,1}^{\text{T}})\circledast \mathbf{e}_{r,3}^\text{R}-(\mathbf{e}_h^{\text{s}}+\mathbf{r}_{r,2}^{\text{T}})\circledast \mathbf{e}_{r,4}^\text{R}+ \\
&\quad\; (\mathbf{e}_h^{\text{t}}+\mathbf{r}_{r,3}^{\text{T}})\circledast \mathbf{e}_{r,1}^\text{R}+(\mathbf{e}_h^{\text{v}}+\mathbf{r}_{r,4}^{\text{T}})\circledast \mathbf{e}_{r,2}^\text{R}], \mathbf{e}_t^{\text{t}}\rangle  \\
&\, +\langle[(\mathbf{e}_h^{\text{j}}+\mathbf{r}_{r,1}^{\text{T}})\circledast \mathbf{e}_{r,4}^\text{R}+(\mathbf{e}_h^{\text{s}}+\mathbf{r}_{r,2}^{\text{T}})\circledast \mathbf{e}_{r,1}^\text{R}- \\
&\quad\; (\mathbf{e}_h^{\text{t}}+\mathbf{r}_{r,3}^{\text{T}})\circledast \mathbf{e}_{r,2}^\text{R}+(\mathbf{e}_h^{\text{v}}+\mathbf{r}_{r,4}^{\text{T}})\circledast \mathbf{e}_{r,3}^\text{R}],\mathbf{e}_t^{\text{v}}\rangle\\
& =[(\mathbf{e}_h^{\text{j}}+\mathbf{r}_{r,1}^{\text{T}})\circledast \mathbf{e}_{r,1}^\text{R}]\cdot (\mathbf{e}_t^{\text{j}})^{\top}-[(\mathbf{e}_h^{\text{s}}+\mathbf{r}_{r,2}^{\text{T}})\circledast \mathbf{e}_{r,2}^\text{R}]\cdot (\mathbf{e}_t^{\text{j}})^{\top}- \\
&\quad\; [(\mathbf{e}_h^{\text{t}}+\mathbf{r}_{r,3}^{\text{T}})\circledast \mathbf{e}_{r,3}^\text{R}]\cdot (\mathbf{e}_t^{\text{j}})^{\top}-[(\mathbf{e}_h^{\text{v}}+\mathbf{r}_{r,4}^{\text{T}})\circledast \mathbf{e}_{r,4}^\text{R}]\cdot (\mathbf{e}_t^{\text{j}})^{\top} \\
&\, +[(\mathbf{e}_h^{\text{j}}+\mathbf{r}_{r,1}^{\text{T}})\circledast \mathbf{e}_{r,2}^\text{R}]\cdot(\mathbf{e}_t^{\text{s}})^{\top}+[(\mathbf{e}_h^{\text{s}}+\mathbf{r}_{r,2}^{\text{T}})\circledast \mathbf{e}_{r,3}^\text{R}]\cdot(\mathbf{e}_t^{\text{s}})^{\top}+ \\
&\quad\; [(\mathbf{e}_h^{\text{t}}+\mathbf{r}_{r,3}^{\text{T}})\circledast \mathbf{e}_{r,4}^\text{R}]\cdot(\mathbf{e}_t^{\text{s}})^{\top}-[(\mathbf{e}_h^{\text{v}}+\mathbf{r}_{r,4}^{\text{T}})\circledast \mathbf{e}_{r,1}^\text{R}]\cdot(\mathbf{e}_t^{\text{s}})^{\top}  \\
&\, +[(\mathbf{e}_h^{\text{j}}+\mathbf{r}_{r,1}^{\text{T}})\circledast \mathbf{e}_{r,3}^\text{R}]\cdot(\mathbf{e}_t^{\text{t}})^{\top}-[(\mathbf{e}_h^{\text{s}}+\mathbf{r}_{r,2}^{\text{T}})\circledast \mathbf{e}_{r,4}^\text{R}]\cdot(\mathbf{e}_t^{\text{t}})^{\top}+ \\
&\quad\; [(\mathbf{e}_h^{\text{t}}+\mathbf{r}_{r,3}^{\text{T}})\circledast \mathbf{e}_{r,1}^\text{R}]\cdot(\mathbf{e}_t^{\text{t}})^{\top}+[(\mathbf{e}_h^{\text{v}}+\mathbf{r}_{r,4}^{\text{T}})\circledast \mathbf{e}_{r,2}^\text{R}]\cdot(\mathbf{e}_t^{\text{t}})^{\top}  \\
&\, +[(\mathbf{e}_h^{\text{j}}+\mathbf{r}_{r,1}^{\text{T}})\circledast \mathbf{e}_{r,4}^\text{R}]\cdot(\mathbf{e}_t^{\text{v}})^{\top}+[(\mathbf{e}_h^{\text{s}}+\mathbf{r}_{r,2}^{\text{T}})\circledast \mathbf{e}_{r,1}^\text{R}]\cdot(\mathbf{e}_t^{\text{v}})^{\top}- \\
&\quad\; [(\mathbf{e}_h^{\text{t}}+\mathbf{r}_{r,3}^{\text{T}})\circledast \mathbf{e}_{r,2}^\text{R}]\cdot(\mathbf{e}_t^{\text{v}})^{\top}+[(\mathbf{e}_h^{\text{v}}+\mathbf{r}_{r,4}^{\text{T}})\circledast \mathbf{e}_{r,3}^\text{R}]\cdot(\mathbf{e}_t^{\text{v}})^{\top}\\
&=\sum_{m}^{|\mathcal{M}|} \sum_{m'}^{|\mathcal{M}|} \langle  \mathcal{R}_{imm'} (\mathbf{e}^m_h),\mathbf{e}^{m'}_t \rangle.
\end{aligned}$}
\end{equation}

\noindent where $\mathcal{R}_{imm'}$ represents the biquaternion algebra translation and rotation transformation between modality $m$ and $m'$, as intuitively shown in Figure 3. Based on the expanded formulation above, we observe that the biquaternion-based score function can be expressed as a linear combination of all pair-wise modality-specific score functions. Furthermore, these score functions independently characterize the translation and rotation of relationships. As a result, M-Hyper is capable of capturing all pairwise semantic relationships $m\xrightarrow{}m'$, ensuring no information redundancy or missing modality combinations.

\subsection*{B Hyperparameter Analysis}
We conducted an analysis of the hyperparameters involved in M-Hyper, with the results presented in Figure~\ref{hyperparameter}. It can be observed that the noise ratio $\beta$ and regularization factor $\lambda$ can improve the model performance within a certain range. However, excessive weights for these parameters negatively impact the model by introducing interference. Additionally, we investigated the model dimension $d$, and the results indicate that insufficient model dimensions fail to adequately capture the characteristics of the data, while overly large dimensions (e.g., $d\geq 512$) do not consistently enhance the model performance.

\begin{figure}
\centering
\includegraphics[scale=0.58]{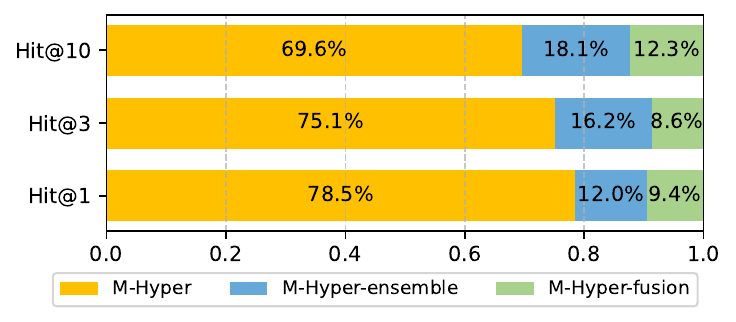}
\vspace{-2mm}
\caption{Comparison of Paradigm Proportions Achieving Optimal Performance Across Relations.} \label{morecase}
\vspace{-3mm}
\end{figure}

\subsection*{C More Case Analysis Between Paradigms}

\begin{table}
\resizebox{0.48\textwidth}{!}{
\centering
\begin{tabular}{llccccc}
\toprule
Relation                &\#Num & IMF    & AdaMF  & MyGO & MoMoK & M-Hyper \\ \midrule
type$^\ddagger$                    & 209   & 37.62 & 34.53 &\underline{50.68} &40.91 & \textbf{52.26}      \\
country$^\ddagger$                 & 352   & 34.97 & 34.42 &\textbf{48.97} &39.16 & \underline{47.34}    \\
language$^\ddagger$                & 82    & 39.69 & 35.75 &\underline{45.99} &42.22 & \textbf{50.73}   \\
time\_Zone$^\ddagger$                & 125   & 34.21 & 35.08 &\textbf{59.71} &37.98 & \underline{54.22}    \\
spouse$^\Diamond$                  & 48    & 34.52 & 28.12 &\underline{55.85} &36.23 & \textbf{65.05}    \\
different\_From$^\Diamond$           & 43    & 23.20 & 32.21 &\underline{38.04} &30.98 & \textbf{40.67}     \\
is\_Part\_Of$^\dagger$                & 183   & 32.98 & 30.05 &\underline{72.16} &39.44 & \textbf{72.77}   \\
company$^\S$                 & 26    & 29.62 & 34.46 &\underline{67.15} &39.44 & \textbf{83.19}   \\
music\_Composer$^\dagger$	&151	&37.61	&32.23	&\underline{47.62}  &42.20 &\textbf{55.69}	 \\
associated\_Band$^\S$ & 255   & 40.67 & 33.09 & \underline{80.76} &43.42 &\textbf{87.17}     \\ \bottomrule 
\end{tabular}}
\vspace{-1mm}
\caption{Results of MRR per relation on DB15K. We mark 1-to-N$^{\dagger}$, N-to-1$^{\ddagger}$, N-to-N$^\S$, and symmetric$^\Diamond$ relations.}
\vspace{-2mm}
\label{case}
\end{table}

\begin{figure}
\centering
\includegraphics[scale=0.45]{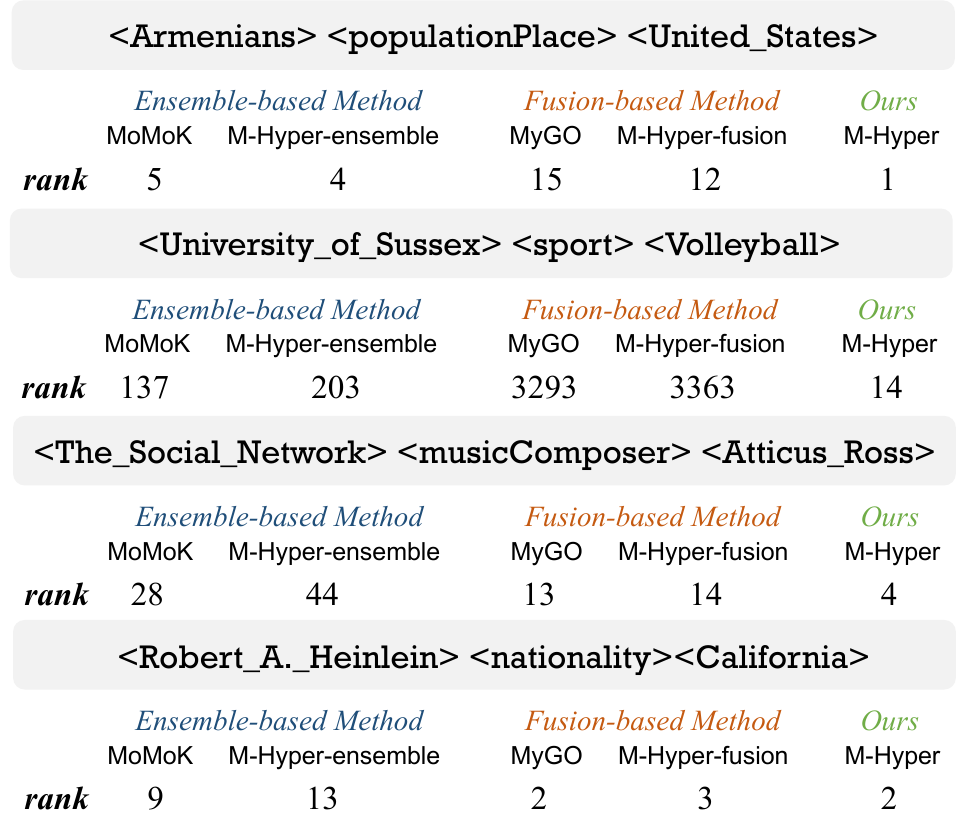}
\vspace{-2mm}
\caption{Intuitive cases show the superiority of M-Hyper.} \label{triplecase}
\end{figure}

\begin{table*}[t]
\centering
\resizebox{0.84\textwidth}{!}{
\begin{tabular}{l|cccc|cccc|cccc}
\toprule
\multirow{2}{*}{Model} & \multicolumn{4}{c|}{\textbf{DB15K}} & \multicolumn{4}{c|}{\textbf{MKG-W}} & \multicolumn{4}{c}{\textbf{MKG-Y}} \\
& MRR & H@1 & H@3 & H@10 & MRR & H@1 & H@3 & H@10 & MRR & H@1 & H@3 & H@10 \\
\midrule
BiQUE & 38.34 & 32.38 & 41.48 & 53.23 & 35.01 & 29.42 & 37.01 & 46.49 & 36.74 & 34.82 & 38.25 & 42.16 \\
MyGO+Tucker & 37.72 & 30.08 & 41.26 & 52.21 & 36.10 & 29.78 & 38.54 & 47.75 & 38.44 & 35.01 & 39.84 & 44.19 \\
MyGO+BiQUE & 37.43 & 29.83 & 41.53 & 52.05 & 35.73 & 29.88 & 37.38 & 46.74 & 37.31 & 35.23 & 38.63 & 42.56 \\
MoMoK+Tucker & 39.57 & 32.38 & 43.45 & 54.14 & 35.89 & 30.38 & 37.54 & 46.13 & 37.91 & 35.09 & 39.20 & 43.20 \\
MoMoK+BiQUE & 40.23 & 33.43 & 43.84 & 54.81 & 35.23 & 29.12 & 36.82 & 46.12 & 37.49 & 35.52 & 38.86 & 42.93 \\
\midrule
\textbf{M-Hyper (Ours)}\qquad\quad & \textbf{41.25} & \textbf{33.64} & \textbf{45.01} & \textbf{56.09} & \textbf{37.02} & \textbf{31.24} & \textbf{39.16} & \textbf{48.84} & \textbf{39.46} & \textbf{36.02} & \textbf{40.92} & \textbf{45.22} \\
\bottomrule
\end{tabular}}
\caption{Baselines with the same decoders, and the embedding dimensions of the BiQUE decoders are kept identical.}
\label{table:para0}
\end{table*}

\begin{table*}[ht]
\centering
\resizebox{0.8\textwidth}{!}{
  \begin{tabular}{lcccccc}
    \toprule
    Model &Training Time (s) & MRR & Hit@1 & Memory Usage (1000MB) &Time Usage (s/epoch) & Params (M)\\
    \midrule
    OTKGE   & 3505  & 23.86 & 18.45 & \textbf{2.540}  & 70.1 & 33.2 \\
    MMRNS   & 7650  & 32.68 & 23.01 & 25.582 & 25.5 & \textbf{3.4}  \\
    AdaMF   & 12500 & 32.51 & 21.31 & 10.428 & 12.5 & 80.7 \\
    IMF     & 7600  & 32.25 & 24.20 & 3.980  & 7.6  & 81.0 \\
    MyGO    & 15900 & 37.72 & 30.08 & 18.128 & 10.6 & 23.4 \\
    MoMoK   & 11700 & 39.57 & 32.38 & 5.900  & 9.8  & 80.6 \\ \midrule
    \textbf{M-Hyper} & \textbf{1200}  & \textbf{40.75} & \textbf{33.14} & 2.862  & \textbf{5.8}  & 21.5 \\
    \bottomrule
  \end{tabular}}
\caption{Detailed performance and overhead comparison of M-Hyper at smaller parameter scales.}
\label{table:para1}
\end{table*}

\begin{table*}[htbp]
  \centering
  \resizebox{0.8\textwidth}{!}{
    \begin{tabular}{lcccccccccc}
      \toprule
      \multirow{2}{*}{Model} & \multicolumn{5}{c}{{DB15K}} & \multicolumn{5}{c}{{MKG-W}} \\
      \cmidrule(r){2-6} \cmidrule(l){7-11}
      & Emb. Size & MRR & Hit@1 & Hit@3 & Hit@10 & Emb. Size & MRR & Hit@1 & Hit@3 & Hit@10 \\
      \midrule
      MyGO    & 800 & 37.83 & 30.09 & 41.31 & 52.28 & 800 & 36.16 & 29.85 & 38.53 & 47.79 \\
      MoMoK   & 800 & 39.62 & 32.47 & 43.44 & 54.14 & 800 & 35.87 & 30.42 & 37.58 & 46.18 \\
      \textbf{M-Hyper} & {800} & \textbf{41.21} & \textbf{33.68} & \textbf{45.06} & \textbf{56.14} & {800} & \textbf{37.02} & \textbf{31.27} & \textbf{39.17} & \textbf{48.84} \\
      \midrule
      MyGO    & 100 & 36.98 & 29.14 & 41.09 & 51.30 & 100 & 35.27 & 29.10 & 37.66 & 46.98 \\
      MoMoK   & 100 & 38.64 & 31.97 & 42.60 & 53.68 & 100 & 35.10 & 29.38 & 36.83 & 45.35 \\
      \textbf{M-Hyper} & {100} & \textbf{40.23} & \textbf{32.79} & \textbf{44.38} & \textbf{55.23} & {100} & \textbf{36.21} & \textbf{30.45} & \textbf{38.47} & \textbf{47.98} \\
      \bottomrule
    \end{tabular}
  }\caption{Performance comparison with baselines at the same embedding sizes as M-Hyper on DB15K and MKG-W datasets.}
  \label{table:para2}
\end{table*}

\paragraph{Specific Relation Performance.}To provide a more granular analysis of M-Hyper's advantages, we present the MRR improvements for common relation on DB15K dataset, as shown in Table~\ref{case}. M-Hyper significantly enhances the performance for 1-to-N relations (e.g., \textit{is\_Part\_Of, music\_Composer}), N-to-1 relations (e.g., \textit{country, language, timeZone}), and N-to-N relations (e.g., \textit{company, associated\_Band}). These are challenging for translation-based methods~\cite{IKRL,MOSE} to address. Additionally, M-Hyper can also achieve at least 6.91\% performance improvement in modeling symmetric relationships (e.g., \textit{spouse, different\_From}), demonstrating stronger geometric representation capabilities. More case analysis are presented in Appendix C.

\paragraph{M-Hyper-fusion and M-Hyper-ensemble.} To further investigate the differences in paradigm shifts, we conduct a more detailed comparison by introducing variations of M-Hyper based on traditional paradigms. Specifically, we keep other modules and the dimension of final embeddings consistent while modifying the score function to create variants: \textit{M-Hyper-fusion} with score function $\phi(h, r, t)=\langle(\mathbf{e}^\text{j}_h+\mathbf{r}_r^\text{T})\circledast\mathbf{r}^\text{R}_r,\mathbf{e}^\text{j}_t\rangle$, and \textit{M-Hyper-ensemble} with score function $\phi(h, r, t)=\sum_m^{|\mathcal{M}|}\langle(\mathbf{e}^m_h+\mathbf{r}_{r,m}^\text{T})\circledast\mathbf{r}^\text{R}_{r,m},\mathbf{e}^m_t\rangle$. Figure~\ref{morecase} illustrates the proportion distribution of different paradigms achieving optimal performance across various relations. It can be observed that M-Hyper achieves the highest proportion in the majority of relationships.

\paragraph{Cases M-Hyper Perform Better.}As shown in~\ref{triplecase}, we present several representative examples of triples under different reasoning requirements. For examples like \textit{(Armenians, populationPlace, United\_States)} and \textit{(University\_of\_Sussex, sport, Volleyball)}, the reasoning results tend to rely more on single-modal features, specifically textual semantic features and analogical reasoning through structural features, respectively. We can find fusion-based methods perform better at preserving the original features, thereby achieving more accurate predictions. In contrast, for relatively long-tail case like \textit{(The\_Social\_Network, musicComposer, Atticus\_Ross)}, and for cases where the answer is relatively sub-optimal like \textit{(Robert\_A.\_Heinlein, nationality, California)}, the model often needs to collaborate across multiple modalities, such as text and structural information, to infer the answer. Therefore, in this problem type, ensemble-based methods are more suitable for such cooperative reasoning scenarios. At the same time, we observe that M-Hyper surpasses both of these approaches and is more adaptable to diverse and flexible reasoning requirements.

\begin{table*}[ht]
\centering
\resizebox{0.62\textwidth}{!}{
\begin{tabular}{c|ccccc|cccc}
\toprule
\multirow{2}{*}{{Dataset}} & \multirow{2}{*}{{$|\mathcal{E}|$}} & \multirow{2}{*}{{$|\mathcal{R}|$}} & \multirow{2}{*}{{\#Train}} & \multirow{2}{*}{{\#Valid}} & \multirow{2}{*}{{\#Test}} & \multicolumn{2}{c}{{image}} & \multicolumn{2}{c}{{Text}}\\
 &  &  &  &  &  & Num & Dim & Num & Dim \\
\midrule
{DB15K}  & 12842 & 279 & 79222 & 9902 & 9904 & 12818 & 4096 & 9078 & 768 \\
{MKG-W}  & 15000 & 169 & 34196 & 4276 & 4274 & 14463 & 383 & 14123 & 384 \\
{MKG-Y}  & 15000 & 28 & 21310 & 2665 & 2663 & 14244 & 383 & 12305 & 384 \\
\bottomrule
\end{tabular}
}
\caption{The statistics of three MMKG benchmarks.}
\vspace{-2mm}
\label{table:dataset}
\end{table*}

\subsection*{D Comparison under Different Parameter Settings}
To demonstrate the effectiveness of M-Hyper and address concerns regarding whether the performance gains are derived from increased parameter count, we present results under three distinct parameter settings:

\paragraph{Baselines with the same decoders.}we compare BiQUE (structure-only) against advanced methods equipped with the biquaternion KGE. To ensure fairness, the embedding dimensions for all baselines using the biquaternion KGE decoder are set to be consistent with M-Hyper. Regarding implementation details: for the fusion-based method (MyGO), we split its final fused representation to serve as the input for the biquaternion KGE; for the ensemble-based method (MoMoK), we split the representations of each modality to perform biquaternion KGE calculations separately. As shown in Table~\ref{table:para0}, M-Hyper consistently outperforms both the unimodal baseline and the BiQUE-enhanced baselines. This confirms that our performance gains stem from the holistic architectural design rather than merely the hypercomplex backbone decoder itself.

\paragraph{Baselines with total training parameters $\geq$ M-Hyper.}As shown in Table~\ref{table:para1}, we provide a detailed comparison of performance and efficiency on the DB15K dataset. It can be observed that compared to most state-of-the-art methods, M-Hyper achieves superior performance even with a significantly smaller number of total training parameters.

\paragraph{Baselines with embedding dimensions equal to M-Hyper.}To ensure a fair comparison, for methods utilizing hypercomplex decoders, we aligned the total dimension of their hypercomplex components with the corresponding real-valued models. We present the comparative results under two embedding size settings (800 and 100) in Table~\ref{table:para2}. The experimental results demonstrate that M-Hyper consistently outperforms other baselines under identical embedding dimensions. This indicates that the performance improvement is not solely attributed to an increase in parameter count.

\subsection*{E Pseudocode of ``Noise-powered Self-distillation''}
As shown in pseudocode~\ref{alg:distillation}, we have provided a detailed description of the process of the "Noise-powered Self-distillation" module.

\begin{algorithm}[t]
\centering
\begin{minipage}{0.9\columnwidth} 
\caption{\small Noise-powered Self-distillation}
\label{alg:distillation}

\renewcommand{\algorithmicrequire}{\textbf{Input:}}
\renewcommand{\algorithmicensure}{\textbf{Output:}}

\begin{algorithmic}[1]
\small 

\REQUIRE Noise rate $\beta$; Relation query $r$; Batch of entity embeddings $\mathcal{E} = \{ \mathbf{e}^m \}_{m \in \{s, v, t\}}$.
\ENSURE Distillation Loss $\mathcal{L}_{distill}$

\STATE Initialize $\tilde{\mathcal{E}}_{student} \leftarrow \emptyset$
\FORALL{$m \in \{\text{structural, visual, textual}\}$}
    \STATE Calculate feature mean $\boldsymbol{\varphi}^m$ and variance $\boldsymbol{\mu}^m$
    \STATE Generate noise: $\tilde{\mathbf{e}}^m \sim \mathcal{N}(\boldsymbol{\varphi}^m, \boldsymbol{\mu}^m)$
    \STATE Sample binary mask $\mathbf{M}$ with probability $\beta$
    \STATE Inject noise: $\mathbf{e}^{m\prime} \leftarrow \mathbf{e}^m + \mathbf{M} \odot \tilde{\mathbf{e}}^m$
    \STATE $\tilde{\mathcal{E}}_{student} \leftarrow \tilde{\mathcal{E}}_{student} \cup \{\mathbf{e}^{m\prime}\}$
\ENDFOR

\STATE Compute $r$-aware weights $w^m$ via Eq. (6)
\STATE Fuse clean embeddings: $\hat{\mathbf{e}}^{\mathrm{j}} \leftarrow \sum w^m \mathbf{e}^m$

\STATE Compute weights and fuse noisy embeddings $\tilde{\mathcal{E}}_{student}$
\STATE Obtain student embedding: $\hat{\mathbf{e}}^{\mathrm{j}\prime} \leftarrow \operatorname{R2MF}(\tilde{\mathcal{E}}_{student}, r)$

\STATE Calculate MSE loss: $\mathcal{L}_{distill} = \| \hat{\mathbf{e}}^{\mathrm{j}} - \hat{\mathbf{e}}^{\mathrm{j}\prime} \|^2$

\RETURN $\mathcal{L}_{distill}$
\end{algorithmic}
\end{minipage}
\end{algorithm}

\subsection*{F Dataset Statistics}
The statistical details of dataset are shown in Table~\ref{table:dataset}.